\documentclass[sigconf]{acmart}
\usepackage{colortbl}
\usepackage{algpseudocode}

\usepackage{amssymb}
\usepackage{tabu}
\usepackage{siunitx}
\usepackage{adjustbox}
\usepackage{tabularx}
\usepackage{subfigure}
\usepackage{array}

\usepackage[]{natbib}

\definecolor{cellco}{RGB}{134, 122, 122}
\AtBeginDocument{%
  \providecommand\BibTeX{{%
    \normalfont B\kern-0.5em{\scshape i\kern-0.25em b}\kern-0.8em\TeX}}}

\setcopyright{acmcopyright}
\copyrightyear{2018}
\acmYear{2018}
\acmDOI{10.1145/1122445.1122456}

\acmConference[Woodstock '18]{Woodstock '18: ACM Symposium on Neural
  Gaze Detection}{June 03--05, 2018}{Woodstock, NY}
\acmBooktitle{Woodstock '18: ACM Symposium on Neural Gaze Detection,
  June 03--05, 2018, Woodstock, NY}
\acmPrice{15.00}
\acmISBN{978-1-4503-XXXX-X/18/06}

\title{TransHash: Transformer-based Hamming Hashing for Efficient Image Retrieval}

\begin{document}



\renewcommand{\shortauthors}{Trovato and Tobin, et al.}

\author{Yongbiao Chen}
\email{chenyongbiao0319@sjtu.edu.cn}
\orcid{1234-5678-9012}
\authornotemark[1]
\affiliation{%
  \institution{Shanghai Jiao Tong University}
  \city{Shanghai}
  \country{China}
}

\author{Sheng Zhang}
\email{zhangshe@usc.edu}
\affiliation{%
  \institution{University of Southern California}
  \streetaddress{800,dongchuan road}
  \city{Los Angeles}
  \country{United States}
}

\author{Fangxin Liu}
\email{liufangxin@sjtu.edu.cn}
\affiliation{%
  \institution{Shanghai Jiao Tong University}
  \city{Shanghai}
  \country{China}
}
\author{Zhigang Chang}
\email{changzig@sjtu.edu.cn}
\orcid{1234-5678-9012}
\authornotemark[1]
\affiliation{%
  \institution{Shanghai Jiao Tong University}
  \city{Shanghai}
  \country{China}
}

\author{Mang Ye}
\email{mangye16@gmail.com}
\affiliation{%
  \institution{Wuhan University}
  \city{Wuhan}
  \country{China}
}

\author{Zhengwei Qi}
\email{qizhwei@sjtu.edu.cn}
\affiliation{%
  \institution{Shanghai Jiao Tong University}
  \city{Shanghai}
  \country{China}
}

\begin{abstract}
Deep hamming hashing has gained growing popularity in approximate nearest neighbour search for large-scale image retrieval. Until now, the deep hashing for the image retrieval community has been dominated by convolutional neural network architectures, e.g. \texttt{Resnet}\cite{he2016deep}. In this paper, inspired by the recent advancements of vision transformers, we present \textbf{Transhash}, a pure transformer-based framework for deep hashing learning.
Concretely, our framework is composed of two major modules: (1) Based on \textit{Vision Transformer} (ViT), we design a siamese vision transformer backbone for image feature extraction. To learn fine-grained features, we innovate a dual-stream feature learning on top of the transformer to learn discriminative global and local features. (2) Besides,  we adopt a Bayesian learning scheme with a dynamically constructed similarity matrix to learn compact binary hash codes. The entire framework is jointly trained in an end-to-end manner.~To the best of our knowledge, this is the first work to tackle deep hashing learning problems without convolutional neural networks (\textit{CNNs}). We perform comprehensive experiments on three widely-studied datasets: \textbf{CIFAR-10}, \textbf{NUSWIDE} and \textbf{IMAGENET}. The experiments have evidenced our superiority against the existing state-of-the-art deep hashing methods. Specifically, we achieve 8.2\%, 2.6\%, 12.7\% performance gains in terms of average \textit{mAP} for different hash bit lengths on three public datasets, respectively. 
\end{abstract}

\ccsdesc[500]{Computer systems organization~Embedded systems}
\ccsdesc[300]{Computer systems organization~Redundancy}
\ccsdesc{Computer systems organization~Robotics}
\ccsdesc[100]{Networks~Network reliability}

\keywords{hamming hashing, deep learning, image retrieval, vision transformer }


\maketitle

\section{Introduction}
\begin{figure}
    \centering
    \includegraphics[width=3.3in]{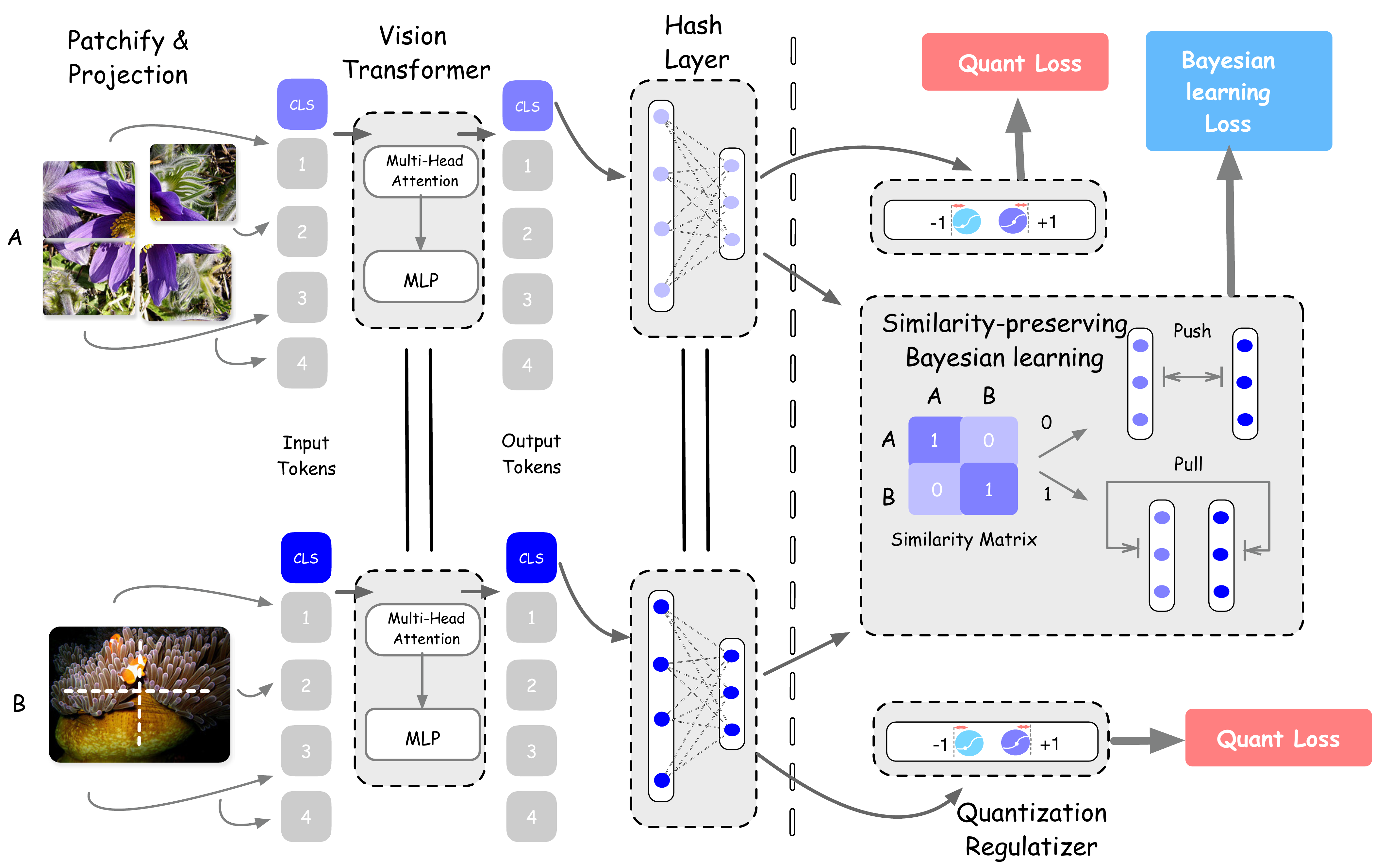}
    \caption{\textbf{The brief architecture of our backbone Siamese vision transformer.} For an image pair (A,B), we cut them into several patches. Then every patch is flattened and projected to a fix-sized embedding with a fully connected layer, resulting in a sequence of embeddings. Subsequently, we add a classification token in the front of each sequence. Then, two sequences are fed into the Siamese transformer architecture. At last, we add a hash layer projecting the feature into B-bit hash vectors. The Bayesian learning module is employed to preserve the similarity in the hashing space for each pair.             }
    \label{fig:subfig}
\end{figure}\par
The past decade has been characterized by the explosive amount of high-dimensional data generated by countless end-users or organizations, resulting in a surge of research attention on accurate and efficient information retrieval methods. Among them, large-scale image retrieval has attained growing traction for its pervasive uses in various scenarios, e.g. recommendation systems, search engines, remote sensing systems. Among all the methods proposed for this challenging task \cite{jegou2010product,ge2013optimized,malkov2018efficient,fu2017fast},~hamming hash-based methods have achieved pronounced successes. It aims to learn a hash function mapping the images in the high-dimensional pixel space into low-dimensional hamming space while preserving their visual similarity in the original pixel space. Scores of works have been introduced. Based on the way they extract features, existing hashing-based works can be divided into two categories, namely, shallow methods and deep learning-based methods.
Shallow methods~\cite{charikar2002similarity,indyk1997locality,weiss2008spectral} learn their hash functions via the hand-crafted visual descriptors (e.g. \textit{GIST}~\cite{oliva2001modeling} ). Nonetheless, the handcrafted features do not guarantee accurate preservation of semantic similarities of raw image pairs, resulting in degraded performances in the subsequent hash function learning process. Deep learning-based~\cite{xia2014supervised,erin2015deep} methods generally achieve significant performance improvements when compared to their shallow counterparts. The common learning paradigm involves two phases. The first phase aims to learn discriminative feature representations with deep convolutional neural networks (CNNs),~e.g. \textit{AlexNet}. The second phase involves designing diversified non  -linear functions to squash the continuous features into binary Hamming codes and devising various ~\cite{liu2016deep,he2018hashing,cao2018deep,cakir2019hashing,fan20deep} losses to preserve the similarity in the raw pixel space. \par
\begin{figure*}
    \centering
    \includegraphics[width=6.7in]{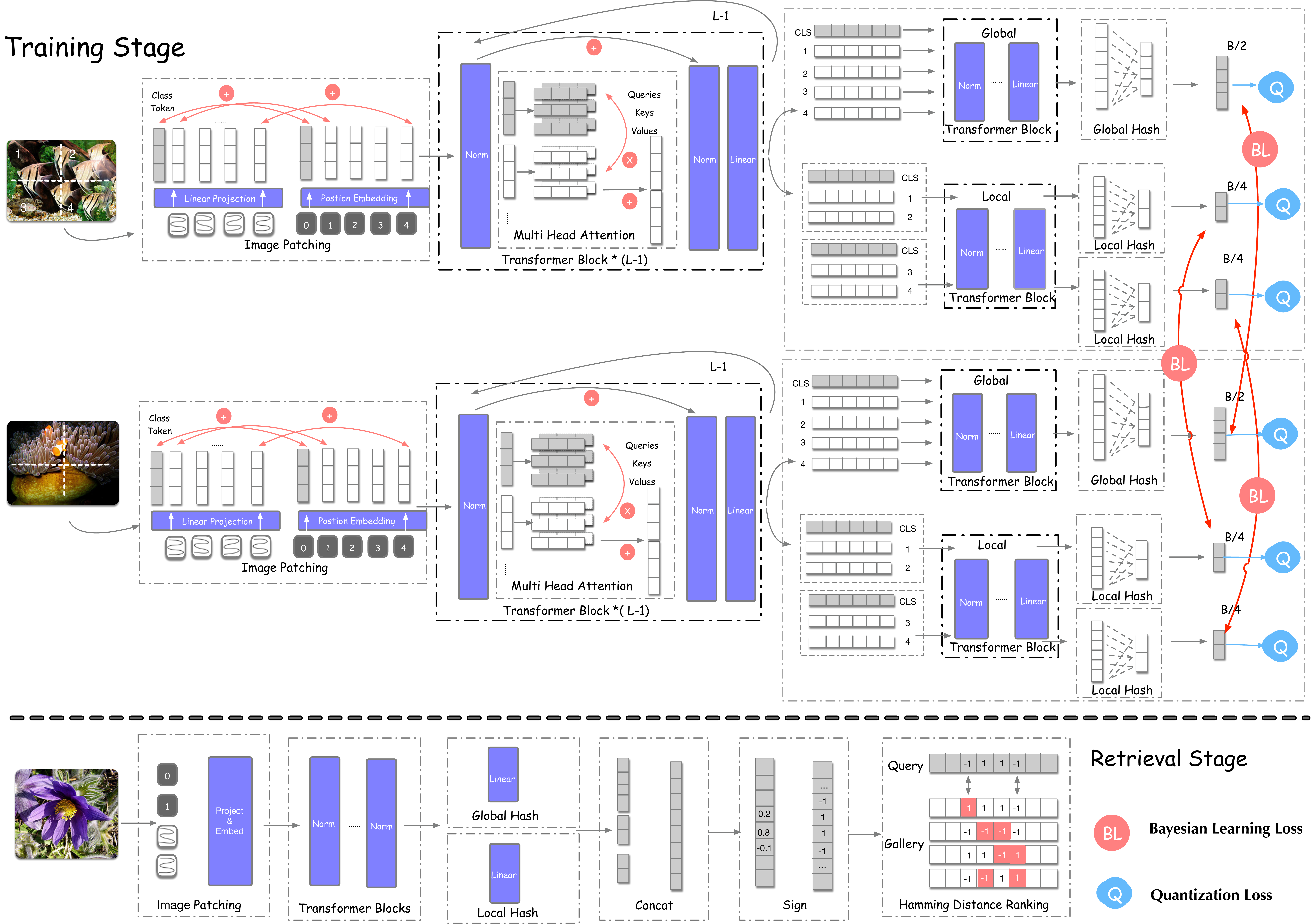}
    \caption{The detailed architecture of the proposed \textbf{TransHash}. The upper part denotes the training stage. Specifically, we follow the same protocols as ViT by feeding the patch embedding together with the position embedding into the transformer encoder. At the last layer of the transformer, we design two parallel transformer blocks: global and local transformer blocks. For the global feature and each local feature, we design a specific hashing layer. In the testing stage, the global and all the local hash vectors are concatenated and quantized into one hash code.}
    \label{fig:mainfig}
\end{figure*}
Recently, transformers~\cite{vaswani2017attention} have demonstrated great successes in natural language processing~\cite{devlin2018bert,brown2020language}. With the advent of the \textit{Vision Transformer}, a variant of transformer tailored for computer vision tasks, transformers have trumped numerous CNN-based methods in various computer vision tasks (e.g. image classification~\cite{dosovitskiy2020image}, object re-identification~\cite{he2021transreid}, and etc). As is shown in Fig.~\ref{fig:subfig},~\textit{Vision transformer} works by first reshaping the input images into a sequence of 2D patches. In the later stage, the 2D patches are transformed into \textit{D} dimensional vectors with a trainable linear projection matrix. Then, a sequence of 1D vectors is fed into the standard transformer architecture to learn a usable feature representation. Inspired by the pronounced performances of \textit{ViT} in other vision tasks, we ponder the possibility of innovating novel deep hashing methods with pure transformers.  \par
In this paper, we build up a novel transformer-based hashing method, dubbed \textbf{Transhash}, which is the very first deep hashing method without adopting a convolutional neural network (CNN) as the backbone architecture. Specifically, targeting pairwise deep hashing learning, we design a Siamese Transformer backbone, which is essentially two identical transformers sharing the same weight parameters\cite{bromley1993signature}. On top of this innovation, inspired by \cite{he2021transreid}, we design a dual-stream feature learning module by changing the last layer of two Siamese transformers to two parallel branches. Concretely,~for the first branch, we learn a global feature representation. In parallel, we reorder the sequence of output features from the second last layer into K groups. The K groups are concatenated with the shared output token and then fed into another transformer layer to generate K local features. The primary merits are stated as follows. Firstly, the model could simultaneously learn fine-grained global and local features with the joint global and local stream design. Secondly, similar to \cite{lai2015simultaneous}, which employs a divide-and-encode module to reduce the redundancy of the learned feature representation, our method could achieve similar effects. Since the final learned representation is a concatenation of the global representation and several local representations, the subsets of the final feature vector are loosely correlated, resulting in increased independence and minimized redundancy. To further preserve the semantic similarity of the image pairs in the feature space, we propose to adopt the Bayesian learning framework to pull close similar pairs and push away dissimilar pairs in the embedding space for all the global and local features.  Finally, since the learned feature representations are continuous in nature, we need to adopt the \textit{sign} function $h = \textit{sign}(f)$ to generate binary hamming hash code in the test stage. However, owing to the sizable gap between the continuous feature representation $f$ and the hash code $h$ after \textit{sign} function, which is officially called the \textit{quantization error}\cite{zhu2016deep}, directly generating hash codes with $sign$ function in the testing stage could only lead to sub-optimal retrieval performances. In an effort to bridge the gap in the training stage, we reformulate the similarity-preserving learning problem as a constrained optimization problem. Concretely, on top of the Bayesian learning module, we add a Cauchy quantization loss~\cite{cao2018deep} to statistically bridge the gap between the continuous feature representation and the binary hash coding. \par
To sum up, we make the following contributions:
\begin{enumerate}
    \item We design a Siamese Transformer backbone based on \textit{ViT} which is two identical vision transformers sharing the same weight parameters.
    \item We innovate a novel two-stream feature learning module by changing the last layer of the transformer into two independent parallel branches. In this fashion, we could learn global and local features at the same time. Meanwhile, as stated before, it could also promote the independence of the learned final hash code vector while reducing bit redundancy.
    \item By further adopting the similarity-preserving Bayesian learning module with a quantization constraint, we build up a novel deep hashing framework for large-scale image retrieval with pure transformer. To the best of our knowledge, this is the very first work for deep learning-based hashing without adopting a convolutional neural network as the backbone.
    \item We conduct comprehensive experiments on three widely-studied datasets-~\textbf{CIFAR-10}, \textbf{NUSWIDE} and \textbf{IMAGENET}. The results show that we outperform all the state-of-the-art method across three datasets by large margins.
   
\end{enumerate}

\section{Related Works}
\subsection{CNNs in Computer Vision}
Convolutional neural network was first introduced in \cite{cnnoriginal} to recognize hand-write numbers. It proposes convolutional kernels to capture the visual context and achieves notable performances. Nonetheless, it was not until the innovation of \textit{AlexNet} \cite{alexnet} that the CNN starts to become the workhorse of almost all the mainstream computer vision tasks, e.g. \textit{Instance Segmentation}~\cite{bolya2019yolact,ren2017end}, \textit{Image Inpainting}~\cite{yeh2017semantic,yu2019free}, \textit{Deep hashing}~\cite{cao2017hashnet,cao2018deep}, \textit{Person Re-identification}~\cite{hermans2017defense,ye2018visible,ye2018hierarchical,chen2020maenet} and etc. To further boost the capability of CNNs, a series of deeper and more effective convolutional neural networks have been proposed, e.g. \textit{VGG}~\cite{vgg}, \textit{GoogleNet}\cite{googlenet}, \textit{ResNet}~\cite{resnet}, \textit{EfficientNet}~\cite{tan2019efficientnet} and etc. While CNNs are still dominant across various computer vision tasks, the recent shift in attention to transformer-based architectures has opened up possibilities to adopt transformers as potent alternatives to convolutional neural networks. Our work is among the first endeavour to replace CNNs with pure transformer-based architectures in traditional computer vision tasks.
\subsection{Transformer in Vision}
The \textit{Transformer} is first proposed in \cite{vaswani2017attention} for sequential data targeting at usage in the field of natural language processing (NLP). Since then, many studies have investigated the effectiveness of Transformer in computer vision tasks by feed to Transformer the sequence of feature maps extracted by CNNs~\cite{girdhar2019video,carion2020end,xie2021segmenting}.  
In 2020, \textbf{Google} proposed \textit{Vision Transformer}~(ViT)~\cite{dosovitskiy2020image}, which applies a pure transformer directly to a sequence of image patches for image classification. Variants of \textit{ViT} have achieved remarkable successes. For instance, \cite{liu2021swin}  proposes a hierarchical vision transformer using shifted windows. \cite{wang2021pyramid}  proposes a pyramid vision transformer tailored for dense prediction. Further, \cite{he2021transreid} proposes the first work of designing a pure-transform based architecture for person re-identification. By utilizing the side information and innovating a novel jigsaw branch, it achieves state-of-the-art across multiple object re-identification datasets. \textit{Vision transformer} is still in its nascent stages. Mounting research attention is being directed to investigate its potential in diversified computer vision tasks.

\subsection{Hashing for Image Retrieval}
Deep hashing for large-scale image retrieval has been drawing growing research attention in recent years ~\cite{indyk1997locality,weiss2008spectral,gong2012iterative,heo2012spherical,jegou2010product,ge2013optimized}. According to the way they extract the features, we could categorize the existing hashing methods into two groups: shallow hashing methods and deep learning based-hash methods. \par
Typical shallow methods are reliant upon the handcraft features to learn a hashing function mapping visual images into binary hash codes. A canonical example is \textbf{LSH} (Locality Sensitive Hash) \cite{indyk1997locality}, which seeks to find a locality-sensitive hash family where the probability of hash collisions for similar objects is much higher than those dissimilar ones. Later, \cite{charikar2002similarity} further proposed another variant of LSH (dubbed SIMHASH) for cosine similarities in Euclidean space. Though these handcrafted feature-based shallow methods achieved success to some extent, when applied to real data where dramatic appearance variation exists, they generally fail to capture the discriminative semantic information, leading to compromised performances. In light of this dilemma, a wealth of deep learning-based hash methods have been proposed~\cite{cao2017hashnet,fan20deep,li2015feature,zhu2016deep,liu2016deep}, for the first time, proposes to learn the features and hash codes in an end-to-end manner. \cite{zhu2016deep} offers a Bayesian learning framework adopting pairwise loss for similarity preserving. \cite{cao2018deep} further suggests substituting the previous probability generation function for neural network output logits with a Cauchy distribution to penalize similar image pairs with hamming distances larger than a threshold. \cite{zhang2019improved} innovates a new similarity matrix targeting multi-label image retrieval.~\cite{fan20deep} further introduces a deep polarized loss for the hamming code generation, obviating the need for an additional quantization loss.


\section{Proposed Method}
In this section, we will elaborate on the design of our framework. 

\paragraph{Problem Formulation}
Suppose we have a training set $T = \{I_i\}_{i=1}^{NT}$ containing $NT$ training images and the corresponding label set $Y = \{y_i\}_{i=1}^{NT}$. For all the pairs of images in the training set, we can construct a similarity matrix $\mathbf{S}$ where $s_{ij} = 1$ if $I_i$ and $I_j$ are from the same class and $s_{ij} = 0$ otherwise. The goal of deep hash for image retrieval is to learn a non-linear hash function $\mathcal{H}: \mathbf{I} \mapsto \{0,1\}^B $ which encodes each input image $I_i$ into a binary hash vector $h_i$  with $B$ bits while preserving the similarity information conveyed in $\mathbf{S}$. That is to say, the Hamming distance between $h_i$ and $h_j$ should be small if $s_{ij} = 1$ and large otherwise.

\subsection{Siamese Vision Transformer Architecture}
\label{sec:siamese}
An overview of our architecture is illustrated in Fig.~\ref{fig:mainfig}. For an image pair $(I_i,I_j)$ with size $H \times W \times 3$, we cut them into identical small patches of patch size $P \times P \times 3$. In doing so, we obtain $N$ patches in total, where $N = H \times W / P^2$. Note that $N$ is also the effective input sequence length for the transformer. 
\\

\paragraph{Patch embeddings} 
For each image patch of $P \times P \times 3$, we flatten it into a vector of size $P^2 \times 3$. Subsequently, similar to \textit{ViT}, we embed every vector into $D$ dimensions with a trainable linear projection (fully connected layer), resulting in a sequence $\{x_p^k\} \in \mathbb{R}^{D}, k \in [1, N]$. We further prepend a learnable embedding $x_{class}$ to $x$, whose state at the end of the output layer serves as the image representation. In this way, we obtain the final embedding $X_p \in \mathbb{R}^{(N+1) \times D}$
\paragraph{Position embeddings}
Positional embedding is adopted to encode the position information of the patch embedding, which is important for the transformer to learn the spatial information of each patch inside the original image. We follow the standard procedure in \textit{ViT} by adding trainable 1D position embedding for every vector in the sequence. Thus, the input for the transformer encoder $z_0$ is stated as follows:
\begin{equation}
    z_0 = X_p + E_{pos} = [x_{class}; x_p^1 , ... , x_p^N ] + E_{pos} 
\end{equation}
\paragraph{Self-attention encoder}  
The transformer encoder consists of $L-1$ blocks, each block containing a multi-headed self-attention layer (\textbf{MSA}) and \textbf{MLP} layer. A layer norm (\textbf{LN}) is applied before each layer while residual connections are applied after each layer, as shown in Fig.~\ref{fig:mainfig}. The computation of a block $\mathcal{F}_{block}$ could be formulated as: 

\begin{equation}
  \begin{split}
      z_{l} &= \mathcal{F}_{msa}(\mathcal{F}_{ln}(z_{l-1})) + z_{l-1}   \\
    z_{l} &= \mathcal{F}_{mlp}(\mathcal{F}_{ln}(z_{l} )) + z_{l} \\
    \text{where} \\ l & = 1 ... (L-1)
  \end{split}
\end{equation}

\paragraph{Dual-stream feature learning} 
After the before-mentioned self-attention encoder, we get the hidden features which are denoted as $Z_{L-1} = [z_{L-1}^0;z_{L-1}^1,z_{L-1}^2, ... ,z_{L-1}^N]$. Note that, as stated before, $z_{L-1}^0$ is the hidden feature for the prepended learnable embedding $x_{class}$. Inspired by \cite{he2021transreid},~we design two parallel branches, the global branch $\mathcal{F}^g_{block}$ and the local branch $\mathcal{F}^l_{block}$. For the global branch, it serves as a standard transformer block encoding $Z_{L-1}$ into $Z_{L} = [f_g;z_L^1,z_L^2, ... ,z_L^  N]$, where $f_g$ is regarded as the global feature representation. For the local branch, we split $Z_{L-1}$ into $K$ groups and prepend the shared token $z_{L-1}^0$ before each group. In this fashion, K feature groups are derived which are denoted as$ \{[z_{L-1}^0;z_{L-1}^1,...,z_{L-1}^{N/K}],$ $ [z_{L-1}^0;z_{L-1}^{N/K+1},...,z_{L-1}^{2\times N/K}], [z_{L-1}^0;z_{L-1}^{N-N/K+1}$ $,...,z_{L-1}^N]\}$. Then, we feed $K$ features groups into $\mathcal{F}^l_{block}$ to learn $K$ local features $\{f_l^1,f_l^2,...,f_l^K\}$. 
\paragraph{Hash layer}
 In an effort to learn compact hash codes, we further design several hash layers projecting every feature vector into different bit sized hash vectors. Concretely, suppose the hash bit length in the retrieval stage is $B$ for each image, then, for the global feature vector of embedding size $M$, we obtain a $B/2$ bit global hash vector through
 \begin{equation}
     h_g = \mathcal{F}_h^g(f_g) = f_g W^T + b 
 \end{equation}
 where $W$ is a weight parameter matrix of size $(B/2,M)$ and b is the bias parameter of size $(B/2,)$. In a similar fashion, for each local feature $f_l \in \{f_l^1,...,f_l^K\}$, we design a specific fully connected layer with $B/(2*K)$ output logits, resulting in $K$ hash vectors $\{h_l^1,...,h_l^K\}$. \par
 In this way, for a image pair $(I_i,I_j)$, the siamese model outputs two sets of hash vectors: $\{\{h_g\}^i,\{h_l^1\}^i,...,\{h_l^K\}^i\}$ and $\{\{h_g\}^j,\{h_l^1\}^j,$ $..., \{h_l^K\}^j\}$, respectively.  
\subsection{Similarity-preserving Bayesian Learning}
\label{sec:similarity}
In this paper, we propose to adopt a Bayesian learning framework for similarity-preserving deep hashing learning. Given training images $(I_i,I_j,s_{ij}): s_{ij} \in \textbf{S}$, where $s_{ij} =1 $ if $I_i$ and $I_j$ are from the same class and $0$ otherwise, we can formulate the logarithm Maximum a Posteriori (\textbf{MAP}) estimation of the hash codes $\boldsymbol{H} = \{ h_1, h_2,...,h_P\}$ for $P$ training points as:
\begin{equation}
\begin{aligned}
\log P(\boldsymbol{H} \mid \mathbf{S}) & \propto \log P(\mathbf{S} \mid \boldsymbol{H}) P(\boldsymbol{H}) \\
&=\sum_{s_{i j} \in \mathbf{S}} w_{i j} \log P\left(s_{i j} \mid \boldsymbol{h}_{i}, \boldsymbol{h}_{j}\right)+\sum_{i=1}^{NT} \log P\left(\boldsymbol{h}_{i}\right)
\end{aligned}
\label{eq:map}
\end{equation}
where $ P(\mathbf{S} \mid \boldsymbol{H})$ is the weighted likelihood function and $w_{ij}$ is the corresponding weight for each image pair $(I_i,I_i)$. Since the similarity matrix $\mathbf{S}$ could be very sparse in real retrieval scenarios~\cite{cao2017hashnet}, it could lead to the data imbalance problem, resulting in sub-optimal retrieval performances. The weighted likelihood is adopted to tackle this problem by assigning weights to each training pair according to the importance of misclassifying that pair~\cite{dmochowski2010maximum}. To be clear, we set 
\begin{equation}
    w_{i j}=\left\{\begin{array}{ll}
|\mathbf{S}| /\left|\mathbf{S}_{1}\right|, & s_{i j}=1 \\
|\mathbf{S}| /\left|\mathbf{S}_{0}\right|, & s_{i j}=0
\end{array}\right.
\end{equation}
where $\mathbf{S}_{1}=\left\{s_{i j} \in \mathbf{S}: s_{i j}=1\right\}$ is the set of similar pairs, $\mathbf{S}_{0}=\left\{s_{i j} \in \mathbf{S}: s_{i j}=0\right\}$ being the set of dissimilar pairs. For an pair $h_i,h_j$, $P\left(s_{i j} \mid \boldsymbol{h}_{i}, \boldsymbol{h}_{j}\right)$ is the conditional probability function of $s_{ij}$ given a pair of hash codes $h_i$ and $h_j$. Since the $s_{ij}$ only takes two values $0$ and $1$, it is natural to define  $P\left(s_{i j} \mid \boldsymbol{h}_{i}, \boldsymbol{h}_{j}\right)$ as a Bernoulli distribution:
\begin{equation}
    \begin{aligned}
P\left(s_{i j} \mid \boldsymbol{h}_{i}, \boldsymbol{h}_{j}\right) &=\left\{\begin{array}{ll}
\sigma\left(\mathcal{D}_H\left(\boldsymbol{h}_{i}, \boldsymbol{h}_{j}\right)\right), & s_{i j}=1 \\
1-\sigma\left(\mathcal{D}_H\left(\boldsymbol{h}_{i}, \boldsymbol{h}_{j}\right)\right), & s_{i j}=0
\end{array}\right.\\
&=\sigma\left(\mathcal{D}_H\left(\boldsymbol{h}_{i}, \boldsymbol{h}_{j}\right)\right)^{s_{i j}}\left(1-\sigma\left(\mathcal{D}_H\left(\boldsymbol{h}_{i}, \boldsymbol{h}_{j}\right)\right)\right)^{1-s_{i j}}
\end{aligned}
\label{eq:beyesian}
\end{equation}
where $\mathcal{D}_H(.)$ is the Hamming distance function and $\sigma$ is a probability function which takes as input a distance of a hash code pair and generate the probability of them from the same class. Note that, since directly optimizing the discrete binary hash code is super challenging, in the training stage, we apply continuous relaxation to the binary constraints $\mathbf{h}_i \in \{-1,1\}^B$ similar to \cite{cao2017hashnet,cao2018deep,zhu2016deep}. Thus, we adopt a surrogate $\mathcal{D}_S$ for $\mathcal{D}_H$ in the continuous space which is formulated as:
\begin{equation}
\begin{aligned}
\mathcal{D}_S\left(\boldsymbol{h}_{i}, \boldsymbol{h}_{j}\right) &=\frac{K}{4}\left\|\frac{\boldsymbol{h}_{i}}{\left\|\boldsymbol{h}_{i}\right\|}-\frac{\boldsymbol{h}_{j}}{\left\|\boldsymbol{h}_{j}\right\|}\right\|_{2}^{2} \\
&=\frac{K}{2}\left(1-\cos \left(\boldsymbol{h}_{i}, \boldsymbol{h}_{j}\right)\right)
\end{aligned}    
\label{eq:prob}
\end{equation}
For the probability function $\sigma$, the most commonly used is the \textit{sigmoid} function. Nevertheless, as stated in \cite{cao2018deep}, the probability of \textit{sigmoid} when the input Hamming distance is much larger than $2$ stays high and only starts to decrease when it approaches $b/2$.~This property makes it hard for the deep hashing method to pull the distance of similar pairs close to a sufficient amount. In light of this dilemma, we propose to adopt \textit{Cauchy} distribution function:
\begin{equation}
    \sigma\left(\mathcal{D}_S\left(\boldsymbol{h}_{i}, \boldsymbol{h}_{j}\right)\right)=\frac{\gamma}{\gamma+\mathcal{D}_S\left(\boldsymbol{h}_{i}, \boldsymbol{h}_{j}\right)}
\label{eq:cauchy}
\end{equation}
where $\gamma$ denotes the scale parameter of the \textit{Cauchy} distribution. The \textit{Cauchy} distribution has a desirable property. The probability of \textit{Cauchy} declines very fast even when the Hamming distance is small, enabling the hashing method to pull the similar images into a small Hamming radius. 
By taking Eq.~\ref{eq:cauchy},~Eq.~\ref{eq:prob}, Eq.~\ref{eq:beyesian} into the \textbf{MAP} estimation in Eq.~\ref{eq:map}, we could derive the optimization objective of similarity-preserving loss as:
\begin{equation}
\begin{aligned}
     L_{s} &= \sum_{s_{i j} \in \mathbb{S}} L_{ce}(\boldsymbol{h}_i,\boldsymbol{h}_j) \\
     &=\sum_{s_{i j} \in \mathbf{S}} w_{i j}\left(s_{i j} \log \frac{\mathcal{D}_s\left(\boldsymbol{h}_{i}, \boldsymbol{h}_{j}\right)}{\gamma}+\log \left(1+\frac{\gamma}{\mathcal{D}_S\left(\boldsymbol{h}_{i}, \boldsymbol{h}_{j}\right)}\right)\right)
\end{aligned}
\label{eq:final}
\end{equation}
From Eq.~\ref{eq:beyesian} and Eq.~\ref{eq:final},~we can observe that $L_{s}$ takes a similar form as logistic regression. By optimizing $L_s$, for a similar pair $(I_i,I_j)$, we are increasing the value of $P(1|\textbf{h}_i,\textbf{h}_j)$, resulting in decreased value of $D_S(\textbf{h}_i,\textbf{h}_j)$ since $\sigma$ is a monotonically decreasing \textit{Cauchy} function.  \\
The quantization constraint to bridge the gap between continuous features and their binary counterparts ($L_Q$) can be derived from the proposed prior $
    P\left(\boldsymbol{h}_{i}\right)=\frac{\gamma}{\gamma+\mathcal{D}_S\left(\left|\boldsymbol{h}_{i}\right|, \mathbf{1}\right)}$
where $\gamma$ is the same scale parameter as Eq.~\ref{eq:cauchy} and $\mathbf{1}$ is a vector of ones. Since we are maximizing $P(H)$ in Eq.~\ref{eq:map}, the quantization loss $L_Q$  is stated as:
\begin{equation}
  L_Q =  \sum_{i=1}^{NT} Q(\boldsymbol{h}_i)=\sum_{i=1}^{NT} \log \left(1+\frac{\mathcal{D}_S\left(\left|\boldsymbol{h}_{i}\right|, \mathbf{1}\right)}{\gamma}\right)
\end{equation}
where $\textbf{1}$ is a vector of ones. By minimizing the quantization loss $Q$ in the training stage, each dimension of the hash vector $\textbf{h}$ is pushed to approximate 1.
\subsection{End to End training}
In this section, we will derive the overall optimization objective of our proposed \textbf{Transhash} method based on Sec.~\ref{sec:siamese} and Sec.~\ref{sec:similarity}. Given  training images in pairs such as  $(I_i,I_j)$, we obtain a pair of continuous hash vector sets $\{\{h_g\}^i,\{h_l^1\}^i,...,\{h_l^K\}^i\}$ and $\{\{h_g\}^j,\{h_l^1\}^j$ $,...,\{h_l^K\}^j\}$ through the siamese vision transformer. Subsequently,~for the local features, we obtain the Bayesian loss and quantization loss as:
\begin{equation}
\begin{aligned}
    L_{B}^{local} = \sum_{s_{i j} \in \mathbf{S} } \sum_{k}^K L_{ce}(\{\textbf{h}_l^k \}^i,\{\textbf{h}_l^k \}^j) \\
    L_{Q}^{local} = \sum_i^{NT} \sum_j^K Q(\{h_l^j\}^i)
\end{aligned}
\end{equation}
where $N$ is the total number of training images, $\mathbf{S}$ represents the similarity matrix, and $K$ denotes the number of local features for each image. In a similar fashion, we could derive the losses for the global features. The overall learning objective for \textbf{Transhash} is formulated as:
\begin{equation}
    \min_{\theta} L_B^{global} + L_B^{local} + \lambda (L_Q^{global} + L_Q^{local})
\end{equation}
where $\theta$ denotes the set of parameters of the framework, and $\lambda$ is the hyper-parameter for controlling the importance of the \textit{Cauchy} quantization loss.

\begin{table*}
\newcolumntype{Y}{>{\centering\arraybackslash}X}
\newlength\mylength
\setlength\mylength{\dimexpr 0.8\textwidth-2\tabcolsep}

    \caption{Mean Average Precision (MAP) of Hamming Ranking for Different Number of Bits on Three Datasets}
    
    \begin{tabularx}{\textwidth}{ XX| YYYY || YYYY ||YYYY } \hline
    \rowcolor{black}  \multicolumn{2}{l|}{ \textcolor{white}{\textbf{Datasets}}} & \multicolumn{4}{c||}{\textcolor{white}{\textbf{CIFAR-10}@54000 }} &\multicolumn{4}{c||}{\textcolor{white}{\textbf{NUSWIDE}@5000}  } &\multicolumn{4}{c|}{\textcolor{white}{\textbf{IMAGENET}@1000}}
       \\ \hline
    \multicolumn{2}{l|}{\textbf{Methods}} & 16 bits &32 bits & 48 bits & 64 bits  & 16 bits &32 bits & 48 bits & 64 bits & 16 bits & 32 bits & 48 bits & 64 bits\\\hline
    
    \multicolumn{2}{l|}{\cellcolor{cellco} \textcolor{white}{\textbf{SH}}~\cite{weiss2008spectral} (NeurIPS)} & - & - & - & - & 0.4058 & 0.4209 & 0.4211 & 0.4104 & 0.2066 & 0.3280 & 0.3951 & 0.4191\\
    \multicolumn{2}{l|}{\cellcolor{cellco} \textcolor{white}{\textbf{ITQ}}\cite{gong2012iterative} (TPAMI)} &  - & - & - & - & 0.5086 & 0.5425 & 0.5580 & 0.5611 & 0.3255 & 0.4620 & 0.5170 & 0.5520 \\
    \multicolumn{2}{l|}{\cellcolor{cellco} \textcolor{white}{\textbf{KSH}}\cite{liu2012supervised} (CVPR)} &  - & - & - & - & 0.3561 & 0.3327 & 0.3124 & 0.3368 & 0.1599 & 0.2976 & 0.3422 & 0.3943 \\
    \multicolumn{2}{l|}{\cellcolor{cellco} \textcolor{white}{\textbf{BRE}}\cite{kulis2009learning} (NeurIPS)} &  - & - & - & - & 0.5027 & 0.5290 & 0.5475 & 0.5546 & 0.0628 & 0.2525 &0.3300 & 0.3578 \\
    
    \hline
    \hline
    \multicolumn{2}{l|}{DSH\cite{liu2016deep12} (CVPR)} & 0.6145 &  0.6815 & 0.6828 & 0.6910 & 0.6338 & 0.6507 & 0.6664 & 0.6856 & 0.4025 & 0.4914 & 0.5254 & 0.5845 \\
  
    \multicolumn{2}{l|}{DHN\cite{zhu2016deep} (AAAI)} & 0.6544 & 0.6711 & 0.6921 & 0.6737 & 0.6471 &  0.6725 & 0.6981 & 0.7027  & 0.4139 &0.4365 &0.4680 & 0.5018 \\ 
  
    \multicolumn{2}{l|}{HashNet\cite{cao2017hashnet} (ICCV) }& 0.5105 & 0.6278 & 0.6631 &0.6826 & 0.6821 & 0.6953 & 0.7193 & 0.7341 & 0.3287 & 0.5789 & 0.6365 & 0.6656 \\
    
    \multicolumn{2}{l|}{DCH\cite{cao2018deep} (CVPR) }& 0.6680 & 0.6936 & 0.6807 & 0.6775 & 0.7036 &0.7178 & 0.7106 & 0.7056 & 0.5868 & 0.5862 &0.5639 & 0.5540 \\
    \multicolumn{2}{l|}{IDHN\cite{zhang2019improved} (TMM) }& 0.5419 & 0.5695 & 0.5895 & 0.5972 &  0.6999 & 0.7149 & 0.7225 & 0.7256 & 0.2583 & 0.3339 & 0.3708 & 0.4037  \\
     \multicolumn{2}{l|}{\cellcolor{cellco} \textcolor{white}{\textbf{DPN}}\cite{fan20deep} (IJCAI) }& 0.825 & 0.838 & 0.830 & 0.829 &  - & - & - & - & 0.684 & 0.740 & 0.756 & 0.756 \\
     \hline
     \multicolumn{2}{l|}{\textbf{TransHash} }&\textcolor{red}{\textbf{0.9075}} & \textcolor{red}{\textbf{0.9108}} & \textcolor{red}{\textbf{0.9141}} & \textcolor{red}{\textbf{0.9166}} & \textcolor{red}{\textbf{0.7263}} & \textcolor{red}{\textbf{0.7393}} & \textcolor{red}{\textbf{0.7532}} & \textcolor{red}{\textbf{0.7488}} & \textcolor{red}{\textbf{0.7852}} & \textcolor{red}{\textbf{0.8733}} & \textcolor{red}{\textbf{0.8932}} & \textcolor{red}{\textbf{0.8921}} \\
     \hline
     \hline
    \end{tabularx}
    \label{table: mainresults}
\end{table*}
\begin{figure*}
    \centering
    \includegraphics[width=7.0in]{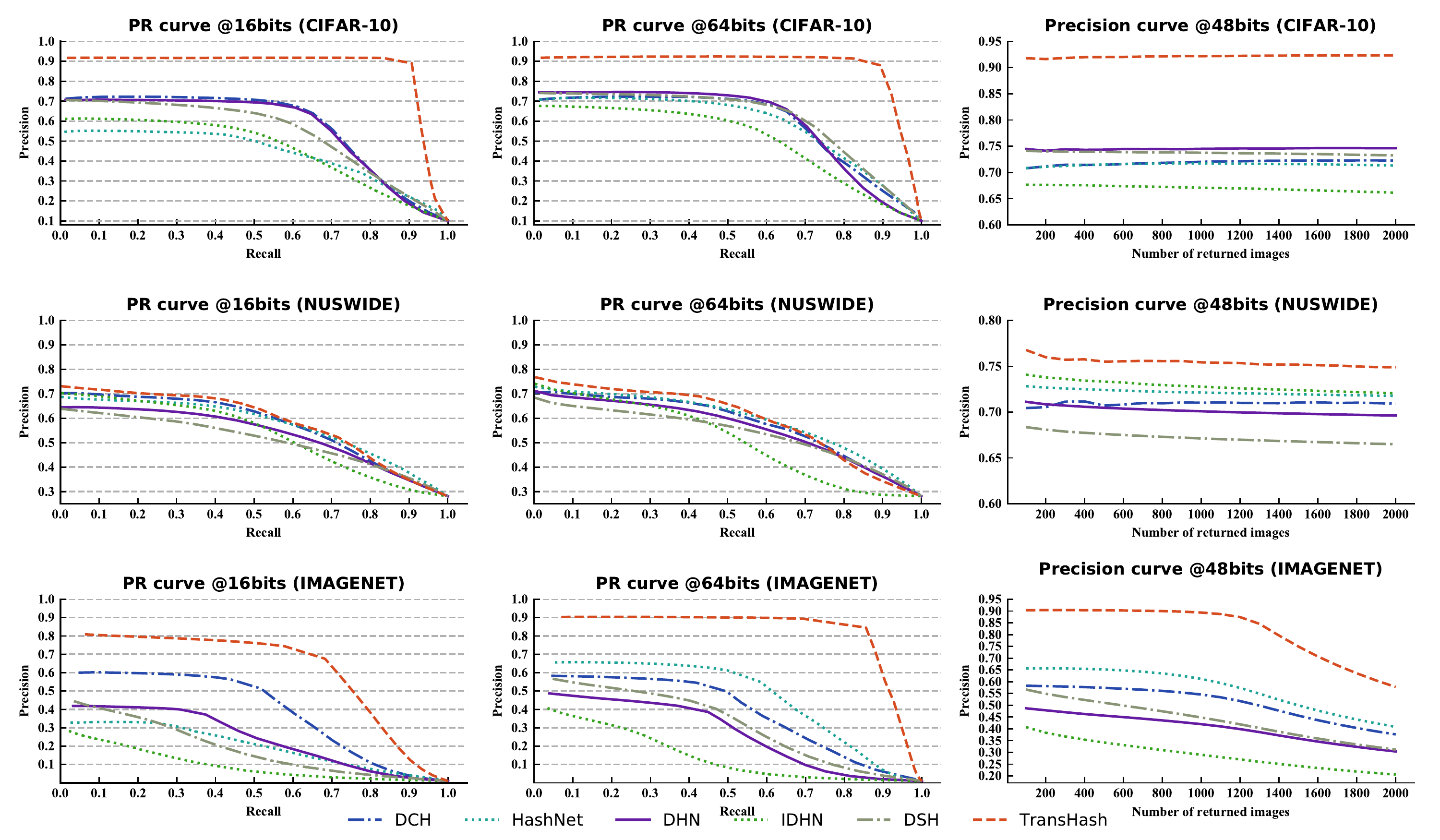}
    \caption{The experimental results of \textbf{TransHash} and other competing methods on three datasets}
     \label{fig:maindata}
\end{figure*}
\subsection{Retrieval Process}
In this section, we will elaborate on how to perform efficient image retrieval given a well-trained model. Generally, we are given a training image set $\textbf{Q}$ and a gallery image set $\textbf{G}$. For an image $I_i^q$ in $\textbf{Q}$, we feed it through the backbone transformer, and obtain a set of hash vectors $\{\{h_g\}^i,\{h_l^1\}^i,...,\{h_l^K\}^i\}$. Subsequently. we concatenate the global and local hash vectors and obtain the final hash vector $\textbf{h}_i^q$:
\begin{equation}
    \textbf{h}_i^q = \textit{sign}(\textit{Concat}([\{\{h_g\}^i,\{h_l^1\}^i,...,\{h_l^K\}^i\}]))
\end{equation}
where $\textit{sign}(x)$ is a element-wise thresholding function which return 1 if $x > 0$ and -1 otherwise. And, \textit{Concat} is a function which concatenate the global and local features into a $B$ bit hash vector. In a similar fashion, for all the images in $\textbf{G} = \{I_k^g\}_{k=1}^{N_g}$, we obtain the binary hash codes $\textbf{H}^g = \{h_k^g\}_{k=1}^{N_g}$. Then, we can rank the binary gallery codes $\textbf{H}^g$ through their Hamming distance with respect to the query hash code $\textbf{h}_i^q$. 
\subsection{Implementation Details}
All the images are first resized to $256 \times 256$. For the training images, we adopt standard image augmentation techniques including \textit{random horizontal flipping} and \textit{random cropping} with cropping size $224$. For testing images, we only apply the \textit{center cropping} with cropping size $224$. The batch size is set to $64$. \textit{SGD} optimizer is adopted with a weight decay of $1e-4$. The learning rate is initialized to $3e-2$ with cosine learning rate decay. The number of warmup steps for the scheduler is set to $500$. The patch size is set to $(32,32)$ for the Siamese transformer model, the hidden size to $1024$. The number of heads for the multi-head attention is set to $16$, and the model consists of $24$ blocks in total.
\section{Experimentation}
\subsection{Datasets and Evaluation Protocols}
\paragraph{Datasets.}  We conduct experiments on three widely-studied image retrieval datasets: \textbf{CIFAR-10},~\textbf{NUSWIDE}, and~\textbf{IMAGENET}. \\
\textbf{CIFAR-10}~\cite{krizhevsky2009learning} is a dataset with $60,000$ images from $10$ classes. We follow the standard protocol in \cite{cao2018deep,zhu2016deep}. Specifically, we randomly select 500 images for each class as the training set, resulting in $5,000$ training points. Then, we randomly select 100 images per class as the query set, the rest denoted as the database. \\
\textbf{NUSWIDE}~\cite{chua2009nus} is a widely-studied public web image dataset consisting of $269,648$ images in total. Each image is annotated with some of the $81$ ground-truth categories (concepts). For fair comparisons, we follow similar experimental protocols ~\cite{cao2017hashnet,zhu2016deep} by randomly sampling $5,000$ as the query set, the rest as the database. Subsequently, we randomly sample $10,000$ images from the database as the training set. \\
\textbf{IMAGENET} is a subset of the dataset for Large Scale Visual Recognition Challenge (ISVRC 2015)~\cite{russakovsky2015imagenet}. Specifically, we follow the same protocol as \cite{fan20deep}\cite{cao2017hashnet} by randomly sampling 100 classes and use all the images of these classes in the validation set as the query set. All the images of these classes in the training set are denoted as the database, while 100 images per category are sampled as the training set.
\paragraph{Evaluation Protocols}
We adopt Mean Average Precision (\textbf{mAP}),\\~\textbf{Precison} ~and \textbf{Recall} as the testing metrics. Concretely, we follow a similar fashion as \cite{cao2018deep,cao2017hashnet}.  The \textbf{mAP} is calculated with the top 54,000 returned images for \textbf{CIFAR-10}, 5,000 for \textbf{NUSWIDE} and 1,000 for \textbf{IMAGENET}

\begin{table*}
\newcolumntype{Y}{>{\centering\arraybackslash}X}

\setlength\mylength{\dimexpr 0.8\textwidth-2\tabcolsep}

    \caption{Mean Average Precision (MAP) of Different Variants of TransHash on Three Datasets}
    
    \begin{tabularx}{\textwidth}{ XX| YYYY || YYYY ||YYYY } \hline
    \rowcolor{black}  \multicolumn{2}{l|}{ \textcolor{white}{\textbf{Datasets}}} & \multicolumn{4}{c||}{\textcolor{white}{\textbf{CIFAR-10}@54000 }} &\multicolumn{4}{c||}{\textcolor{white}{\textbf{NUSWIDE}@5000}  } &\multicolumn{4}{c|}{\textcolor{white}{\textbf{IMAGENET}@1000}}
       \\ \hline
    \multicolumn{2}{l|}{\textbf{Methods}} & 16 bits &32 bits & 48 bits & 64 bits  & 16 bits &32 bits & 48 bits & 64 bits & 16 bits & 32 bits & 48 bits & 64 bits\\\hline

     \hline
     \multicolumn{2}{l|}{\textbf{TransHash} }&\textcolor{red}{\textbf{0.9075}} & \textcolor{red}{\textbf{0.9108}} & \textcolor{red}{\textbf{0.9141}} & \textcolor{red}{\textbf{0.9166}} & \textcolor{red}{\textbf{0.7263}} & \textcolor{red}{\textbf{0.7393}} & \textcolor{red}{\textbf{0.7532}} & \textcolor{red}{\textbf{0.7488}} & \textcolor{red}{\textbf{0.7852}} & \textcolor{red}{\textbf{0.8733}} & \textcolor{red}{\textbf{0.8932}} & \textcolor{red}{\textbf{0.8921}} \\
       \multicolumn{2}{l|}{TransHash w/o \textbf{C} }& 0.8406 & 0.8384 & 0.8958 & 0.9062 &  0.7004 & 0.7265 & 0.7336 & 0.7310 & 0.7172 & 0.7808 & 0.8064 & 0.8244  \\
       \multicolumn{2}{l|}{TransHash w/o \textbf{P} }& 0.9029 & 0.9053 & 0.9028 & 0.9014 &  0.7190 & 0.7147 & 0.7339 & 0.7167 & 0.7549 & 0.8485 & 0.8635 & 0.8635  \\
       \multicolumn{2}{l|}{TransHash w/o \textbf{Q} }& 0.8927 & 0.9023 & 0.9048 & 0.9078 & 0.6540 &  0.6821 & 0.6689 & 0.6915 & 0.7451 & 0.8588 & 0.8689 & 0.8758  \\
     \hline
     \hline
    \end{tabularx}
    \label{table: ablationresults}
\end{table*}

\subsection{Comparison with State-of-the-Arts} \par
In this section, we compare the results of our proposed \textbf{TransHash} and the state-of-the-art deep hashing methods. Specifically, the competing methods could be divided into two categories: shallow hashing methods and deep hashing methods. For the shallow hashing methods, we include the most frequently compared methods \textbf{SH}~\cite{weiss2008spectral},~\textbf{ITQ}~\cite{gong2012iterative},~\textbf{KSH}~\cite{liu2012supervised},~and~\textbf{BRE}~\cite{kulis2009learning} for detailed comparisons. For the deep learning-based hashing methods, we 
further include \textbf{DSH}~\cite{liu2016deep12} which is among the very first works targeting at tackling the hashing problem for image retrieval with deep convolutional neural networks. In addition, we incorporate other recent deep hashing methods including ~\textbf{DHN}\cite{zhu2016deep},~\textbf{HashNet}~\cite{cao2017hashnet},~\textbf{IDHN}~\cite{zhang2019improved} and~\textbf{DPN}~\cite{fan20deep}. \par
Note that, for all the non-deep methods and \textbf{DPN}, we directly quote the results from \cite{cao2017hashnet} and \cite{fan20deep}. For the rest of the competing methods,~ we conduct experiments with the open-sourced codes from the original papers.~For fair comparisons, we conform to original protocols for the hyper-parameters and the pre-processing techniques. For example, all the images are resized to $224 \times 224$.   \par
The Mean Average Precision (\textbf{mAP}) results are demonstrated in Tab.~\ref{table: mainresults}.~It is rather evident that our proposed \textbf{TransHash} is a clear winner compared with the shallow hashing methods across three datasets. Specifically, we achieve absolute performance boosts of 19.93\%,~39.69\% in terms of average \textbf{mAP} for \textbf{NUSWIDE} and \textbf{IMAGENET}, respectively. The unsatisfied performances of these non-deep hashing methods could be in part attributed to the fact that these methods could not assist in the discriminative feature learning process, resulting in the generation of sub-optimal hashing codes. Clearly, deep hashing methods exhibit significantly better performances across all the datasets for different hash bit lengths. Still, our method outperforms all the competing methods by large margins. Specifically,~on \textbf{CIFAR-10}, we achieve a \textbf{mAP} of 91.66\% in terms of 64 hash bits, surpassing the state-of-the-art result by 8.8\%. The performance improvement is even more pronounced in \textbf{IMAGENET}. The average \textbf{mAP} for \textbf{TransHash} is 86.10\%, exceeding \textbf{DPN} by 12.7\%. The reasons for the notable performance gains are twofold. First, the siamese architecture and the dual-stream feature learning design could assist in learning more discriminative features. The second reason is that the ratio between the number of similar pairs and dissimilar pairs in \textbf{IMAGENET} is much larger than \textbf{CIFAR-10}, which is also known as the data imbalance problem~\cite{cao2017hashnet}, deteriorating the performance of methods trained on pairwise data~\cite{zhang2019improved,liu2016deep12}. \textbf{TransHash} tackles this problem by dynamically assigning a weight for each pair as is carried out in \cite{cao2017hashnet}. On \textbf{NUSWIDE}, our method also consistently exceeds the competing methods across different hash bit lengths. The performance gains are not as sizable as on \textbf{CIFAR-10} and \textbf{NUSWIDE} mainly because \textbf{TransHash} is not tailored for multi-label image retrieval where each image comprises multiple labels. \par
We further plot the Precision-Recall curves(PR) in terms of 16 and 64 hash bits and Precision curves with respect to different numbers of top returned images. As depicted in Fig.~\ref{fig:maindata},~the performance of \textbf{TransHash}, colored with red, consistently levitates above all the competing methods by large margins for the PR curves. In terms of precision w.r.t numbers of returned images, as shown in the top right pictures in Fig.~\ref{fig:maindata}, \textbf{TransHash} achieves significantly better results against all the methods. The results on \textbf{NUSWIDE} are on the middle of Fig.~\ref{fig:maindata}. \textbf{TransHash} achieves slightly better results for PR@16 bits and PR@64 bits. For the precision w.r.t number of returned images, our method obtains a precision of 76.77\% for 100 returned images, surpassing \textbf{IDHN} by
2.7\%. Pronounced performance gains could also be spotted for \textbf{IMAGENET}. Specifically, for PR curve with 16 bits, \textbf{DCH} obtains the second place while \textbf{HashNet} tops \textbf{DCH} for 48 bits. It is easy to spot that \textbf{TransHash} still exceeds both methods in two testing scenarios with considerable margins. For the precision curve, we achieve performances of 90.35\%, 89.38\% w.r.t 100 and 1000 returned images, exceeding \textbf{HashNet} by 24.73\% and 28.18\%, respectively. The superior results could sufficiently demonstrate the effectiveness of our pure-transformer-based hashing method.
\begin{figure*}
    \centering
    \includegraphics[width=7.0in]{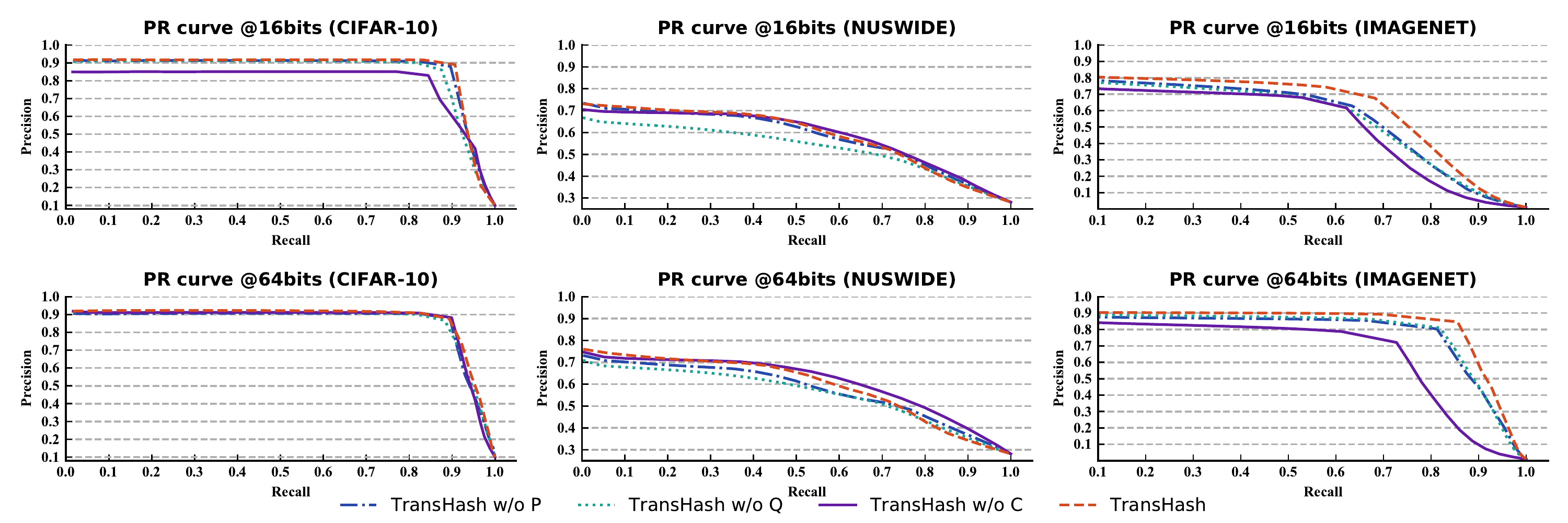}
    \caption{Experimental results of different variants of \textbf{TransHash} on three datasets}
    \label{fig:my_ablation}
\end{figure*}
\subsection{Ablation Studies}
To further analyze the overall design of our proposed method, we conduct a detailed ablation study to demonstrate the effectiveness of each component. Specifically, we investigate three variants of \textbf{TransHash}: 
\begin{enumerate}
    \item \textbf{TransHash w/o P}, a variant without adopting the dual-stream feature learning.
    \item \textbf{TransHash w/o Q}, a variant without the Cauchy quantization loss.
    \item \textbf{TransHash w/o C}, a variant adopting the sigmoid function as the probability function $\sigma$, following the protocols in \cite{zhu2016deep}.
\end{enumerate}

\begin{table}[h]
    \centering
     {
    \begin{tabular}{cc||c||c||c||c||c}
        \hline
        \hline
         \rowcolor{black} \multicolumn{2}{c||}{(\textcolor{white}{Groups \textbf{(K)}})} & \multicolumn{1}{c||}{\textcolor{white}{\textbf{2}}} &\multicolumn{1}{c||}{\textcolor{white}{\textbf{3} }} & \multicolumn{1}{c||}{\textcolor{white}{\textbf{4} }} & \multicolumn{1}{c||}{\textcolor{white}{\textbf{5} } } & \multicolumn{1}{c||}{\textcolor{white}{\textbf{6} }} \\
        \hline\hline
        \multicolumn{2}{c||}{16 bits} & 0.9075 & - & - & - & -  \\
        \multicolumn{2}{c||}{32 bits} & 0.9108 & 0.9013 & - & - & -  \\
        \multicolumn{2}{c||}{48 bits} & 0.9141 & 0.9017 & 0.9187 & 0.9107 & 0.9143  \\
        \multicolumn{2}{c||}{64 bits} & 0.9166 & 0.9103 & 0.9057 & 0.9062 & 0.8994  \\
        \hline
    \end{tabular} }
    \caption{Analysis of the effects of \textbf{K} on \textbf{CIFAR-10}.~Note that $-$ denotes when K equals a certain number,~the model fails to converge as illustrated in the empirical analysis.}
    \label{teb:ablation}
\end{table}

As shown in Tab.~\ref{table: ablationresults} and Fig.~\ref{fig:my_ablation}, when then Cauchy quantization loss is removed (\textbf{TransHash w/o Q}), we experience notable performance declines in \textbf{NUSWIDE} and \textbf{IMAGENET}, from  74.88\% to 69.15\% and 89.21\% to 87.58 \% for 64 hash bits, respectively. When the model is deprived of Cauchy distribution (\textbf{TransHash w/o C}), which is similar to \cite{zhu2016deep}, we can see that the performance decreases sharply. Specifically, on \textbf{IMAGENET}, it experiences a conspicuous performance drop by an average of 5.55\% \textbf{mAP}. We also note that the drop for shorter hash codes is more severe than longer hash codes. The primary reason is that according to \cite{cao2018deep}, the Cauchy distribution could effectively pull close similar pairs into a small Hamming radius, giving it an edge when the hash code length is short. \par
More importantly, to test the effectiveness of the proposed dual-stream feature learning, we also include the performances of the Siamese model with the solo global feature learning module. As depicted in Fig.~\ref{table: ablationresults}, \textbf{TransHash w/o P} consistently underperform the model with dual-feature learning design.  On \textbf{NUSWIDE} and \textbf{IMAGENET}, the average decline is 2.08\% and 2.83\%, respectively. The above experimental experiments have evidenced the effectiveness of the design of our pure transformer-based hashing framework. Since the hyper-parameter $K$, which controls how many groups we will divide our local features into, is rather important in our design, we further provide an ablation study on the sensitivity of $K$ for various hash bits on \textbf{CIFAR-10}. Note that if the length of the final hash code vector is 16 and $K$ equals 2, then the global feature is responsible for learning the first 8 bit and each local feature vector for the latter 4 bits.

\paragraph{Empirical analysis of \textbf{K}}
As depicted in Tab.~\ref{teb:ablation}, generally, the performance is not very sensitive to $K$. Also, we observe that when the local feature vector is responsible for generating less than 4 bits, the model will fail to converge. In light of the above observations, we empirically set the $K$ to $2$ across four different hash bit lengths. 

\section{Conclusion}
In this paper, we have proposed a novel pure transformer-based deep hashing framework (\textbf{TransHash}) to tackle the challenging large-scale image retrieval problem. Specifically, we innovate a novel Siamese transformer architecture for extracting robust image features with pairwise similarity learning. On top of that, in an attempt to learn more fine-grained features, we propose to add a dual-stream feature learning module to learn global and local features simultaneously. A well-specified Bayesian learning framework is adopted on top of all the pairwise features for similarity-preserving learning. The overall framework is optimized in an end-to-end fashion. We conduct extensive experiments and demonstrate that \textbf{TransHash} yields notable performance gains compared to the state-of-the-art deep hashing methods on \textbf{CIFAR-10}, \textbf{NUSWIDE} and \textbf{IMAGENET} datasets.
\bibliographystyle{ACM-Reference-Format}
\bibliography{sample-base}

\appendix

\end{document}